\pdfoutput=1

\documentclass[11pt]{article}

\usepackage{acl}

\usepackage{times}
\usepackage{latexsym}

\usepackage[T1]{fontenc}

\usepackage[utf8]{inputenc}

\usepackage{microtype}

\usepackage{inconsolata}

\usepackage{enumitem}
\setlist[itemize]{noitemsep, topsep=0pt}

\usepackage{amsmath, amsfonts, amssymb, amsbsy, mathtools, nccmath, bm}

\usepackage{graphicx}
\usepackage{subcaption}
\usepackage{afterpage}

\usepackage{multirow}
\usepackage{booktabs}
\usepackage{dblfloatfix}

\usepackage{dsfont}

\makeatletter
  \newcommand\scripty{\@setfontsize\scripty{6pt}{7}}
\makeatother

\usepackage{hyperref}
\definecolor{darkblue}{rgb}{0.0, 0.0, 0.45}
\hypersetup{colorlinks=true,linkcolor=darkblue,citecolor=darkblue,filecolor=darkblue,urlcolor=darkblue,pagebackref=true,breaklinks=true,bookmarks=true}

\usepackage[multiple]{footmisc}
\makeatletter
\renewcommand\@makefntext[1]%
    {\makebox[4pt][r]{\textsuperscript{\@thefnmark}\,}#1}
\makeatother

\usepackage{xcolor}
\providecommand{\ned}[1]{#1}

\usepackage{thanks-nostar}

\title{The case for evaluating multimodal translation models on text datasets}

\author{Vipin Vijayan$^{\text{1}}$, Braeden Bowen$^{\text{1}}$,
Scott Grigsby$^{\text{1}}$, Timothy Anderson$^{\text{2}}$, Jeremy Gwinnup$^{\text{2}}$\\
$^{\text{1}}$PAR Government Systems Corporation, $^{\text{2}}$Air Force Research Laboratory\\
\texttt{\small \{vipin\_vijayan, braeden\_bowen, scott\_grigsby\}@partech.com},\\\texttt{\small \{timothy.anderson.20, jeremy.gwinnup.1\}@us.af.mil}}

\begin{document}
\maketitle
\begin{abstract}

A good evaluation framework should evaluate multimodal machine translation (MMT) models by measuring 1) their use of visual information to aid in the translation task and 2) their ability to translate complex sentences such as done for text-only machine translation.
However, most current work in MMT is evaluated against the Multi30k testing sets, which do not measure these properties.
Namely, the use of visual information by the MMT model cannot be shown directly from the Multi30k test set results and the sentences in Multi30k are are image captions, i.e., short, descriptive sentences, as opposed to complex sentences that typical text-only machine translation models are evaluated against.

Therefore, we propose that MMT models be evaluated using 1) the CoMMuTE evaluation framework, which measures the use of visual information by MMT models, 2) the text-only WMT news translation task test sets, which evaluates translation performance against complex sentences, and 3) the Multi30k test sets, for measuring MMT model performance against a real MMT dataset.
Finally, we evaluate recent MMT models trained solely against the Multi30k dataset against our proposed evaluation framework and demonstrate the dramatic drop performance against text-only testing sets compared to recent text-only MT models.

\end{abstract}

\section{Introduction}
\label{sec:intro}

Multimodal machine translation (MMT) is the problem of automatically translating text from one language to another with the aid of additional modalities such as image, video, audio. The hypothesis that contextually relevant images would help resolve ambiguities or missing information in machine translation (MT) is persuasive. For example, the word bank in English can mean financial institutions or river banks, which typically are unambiguously different words in other languages. A practical use for MMT is improved translations of closed captioning or subtitles in videos.

Much work in MMT focus on the Multi30k dataset \citep{elliott_multi30k_2016}, a dataset comprising 30,014 image captions and corresponding translations in different languages. Introduced alongside the WMT 2016 Multimodal Translation task, the dataset has sparked a series of papers in multimodal machine translation \citep{huang_attention-based_2016, calixto_incorporating_2017, caglayan_lium-cvc_2017, yao_multimodal_2020, yin_novel_2020, wu_good_2021, li_valhalla_2022}.

\subsection{Evaluating MMT models}
\label{sec:intro_eval}

MMT models are typically evaluated against the Multi30k test sets, which also comprise of image captions and corresponding translations.
It has been suggested before that the Multi30k test sets are not adequate for evaluating MMT models since 1) many MMT models do not make use of the visual context \citep{caglayan_probing_2019} and reported performance improvements reported were due to the regularization effect that come from treating the multimodal data as random noise \citep{wu_good_2021}, 2) the use of visual information by the MMT model cannot be shown directly from the test set results, 3) the image captions are short, descriptive sentences as opposed to complex sentences present in text-only MT evaluation datasets, and 4) most of the captions do not require the image in order to be correctly translated due to the captions being unambiguous \citep{futeral_tackling_2022}.
In this work, we also show that MMT models trained against the Multi30k dataset perform poorly against testing sets used by typical text-only translation models.

There have been previous work in improving the evaluation methodology for MMT. Recently, \citet{futeral_tackling_2022} proposed an evaluation framework, CoMMuTE (Contrastive Multilingual Multimodal Translation Evaluation), where the authors evaluate how well an MMT model can use visual information to perform the translation task.
The challenge in this evaluation for the MMT model is, given a lexically ambiguous sentence in English, and a translation of the ambiguous sentence that fits one of two images, determine from which of the two images the English sentence and its translation fits best. The determination is made by selecting the input text and image pair with lower perplexity of the output. The final CoMMuTE score is calculated using the average accuracy over all of the ambiguous sentence and image pairs. While CoMMuTE is useful for measuring how well the MMT model uses visual information, the sentences present do not represent the complexity present in text-only evaluation sets.

Our desired goal in MMT is to translate complex sentences, such as is done in typical machine translation, with the aid of contextual image information. 
However, most current work in MMT focus on the Multi30k dataset and other small MMT datasets (e.g., VATEX \citep{wang_vatex_2020}), which are typically captions, i.e., short sentences describing the images or videos, along with respective translations.
Since many MMT models train only against the Multi30k datasets, this results in overfitted models that have no practical use in the real world. While ideally there would be a test dataset containing complex sentences with ambiguities that can be resolve with contextual image information, we can use the WMT news translation task test sets \citep{kocmi_wmt_findings_2022}, which contain complex sentences from news media, and are used as a standard in text-only machine translation, along with the CoMMuTE framework, which informs us whether the model uses image information as a proxy for such a testing set.

\subsection{Proposal}

A good evaluation framework should evaluate MMT models by measuring 1) their use of visual information to aid in the translation task and 2) their ability to translate complex sentences such as is done for text-only machine translation.

Thus, to evaluate along both of these axes, we propose that MMT models be evaluated using 1) the CoMMuTE evaluation framework \citet{futeral_tackling_2022}, which measures the use of visual information by MMT models, 2) the text-only WMT news translation task test sets, which evaluates translation performance against complex sentences, and 3) the Multi30k test sets, for measuring MMT model performance against a real MMT dataset.

Performing well against all three of these evaluation measures is a significantly more difficult task than performing well against the Multi30k test sets alone. As we show in our work, simply training the model against the Multi30k dataset will not result in good performance against all three metrics. %

\subsection{Our contributions}

\begin{itemize}
\item We design an evaluation framework that takes into consideration the ability of MMT models to 1) use visual information to aid in the translation task and 2) translate complex sentences like typical text-only MT models.
\item \ned{We show that while MMT models trained solely against the Multi30k training set perform well against the Multi30k test sets, the models perform very poorly against the WMT news translation task test sets.}
\end{itemize}

\section{Related Works}
\label{sec:related}

As discussed in Section \ref{sec:intro_eval}, there has been many works on evaluation of MMT models. Here, we focus on the MMT models and datasets we evaluate in this work.

\citet{wu_good_2021} discovered the improvements achieved by the multimodal models over text-only counterparts are in fact results of the regularization effect since the models learn to ignore the multimodal information. They introduced two interpretable MMT models, Gated Fusion and RMMT, and performed \ned{non-matching} evaluation on their models (by inputting random images instead of contextual images with the source text) and showed that adding model regularization achieves results that are comparable to incorporating visual context.

\citet{futeral_tackling_2022} focused on the challenge of resolving ambiguity in the source text by using contextual images for MMT. They proposed an MMT model that combines the standard MMT objective with a
visually-conditioned masked language modeling
(VMLM) objective \citep{li_visualbert_2020,lu_vilbert_2019,su_vl-bert_2020}. They showed that their model was able to successfully translate ambiguous source text into unambiguous output text while using the input image as context. Importantly, they introduced the CoMMuTE evaluation framework to demonstrate that the model can successfully disambiguate the source text and use vision information to perform the translation task. We use the CoMMuTE framework as part of our proposed evaluation framework.

\section{Methods}
\label{sec:methods}

\subsection{Proposed evaluation framework}
\label{sec:metrics}

As discussed in Section \ref{sec:intro}, we evaluate using 1) the CoMMuTE evaluation framework, 2) the WMT news translation task test sets, and 3) the Multi30k test sets.

\begin{figure}[t]
\centering
\small
\begin{tabular}{*{2}{c}}

\multicolumn{2}{c}{Input: Get away from the float!} \\
\includegraphics[height=0.25\linewidth]{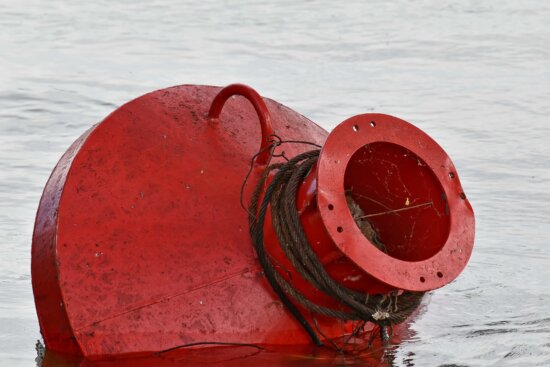} 
 & \includegraphics[height=0.25\linewidth]{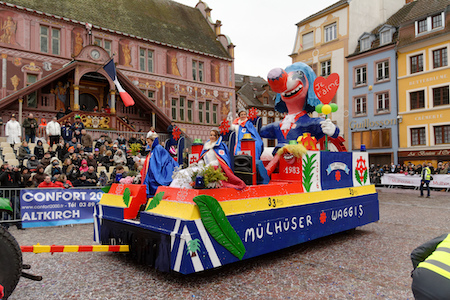} \\ 
Ref: Veg vom Schwimmer! & Ref: Veg vom Festwagen! \\

\end{tabular}
\caption{Example ambiguous English sentence, contextual images, and unambiguous non-English reference sentences from the CoMMuTE test dataset.}
\label{fig:commute}
\end{figure}

\subsubsection{CoMMuTE}
The CoMMuTE framework \citep{futeral_tackling_2022} evaluates how well the MMT model can use visual information to perform the translation task.
The CoMMuTE testing set contains 50 lexically ambiguous English sentences, where each sentence is associated with two images that disambiguate the English sentence into two non-English sentences (Figure \ref{fig:commute}).

For each ambiguous English sentence and associated image pair, the challenge in this evaluation framework for the MMT model is to select which of the two non-English sentences matches up best with the English sentence-image pair.

The determination is made by calculating the perplexity of the (English sentence, image, non-English sentence) triplet using the MMT model. This is done for the two non-English sentences. Then, the non-English sentence with the lower perplexity is selected as the correct translation.
The final CoMMuTE score is calculated using the average accuracy over all of the 100 ambiguous English sentence and image pairs.

\subsubsection{WMT news translation task test sets}

The \ned{WMT news translation task test sets (newstest)} have been used in the WMT machine translation task for many years \citep{kocmi_wmt_findings_2022}. The newstest test sets consists of sentences taken from news media, where the sentences range from two words to 170 words (for the English sentences). Each year, the WMT challenge evaluates translation models against a newstest test set for that particular year and release the past year's translations. In this work, \ned{since the FAIR-WMT19 model that we compare against was trained on newstest test sets from years 2018 and prior}, we evaluate against newstest test sets from years 2019 (1,997 sentences) and 2020 (1,418 sentences) using BLEU4.

\subsubsection{Multi30k}

The Multi30k dataset was introduced alongside the WMT 2016 multimodal translation task, which introduced the task of MMT. The Multi30k dataset contains 29,000 training images and 1,014 validation images with parallel sentences in English and German. The Multi30k images are from the Flickr30k dataset \citep{young_image_2014} with the German translations created by professional translators from the original English image descriptions.

We evaluate against the Multi30k Test2016 test set (1000 examples) and the Test2017 test set (1000 examples) using BLEU4.

\subsection{MMT models}
\label{sec:approach}

We evaluate the Gated Fusion and RMMT multimodal translation models proposed by \citet{wu_good_2021} that performs well against the Multi30k test sets, and compare the results against the FAIR-WMT19 \citep{ng_facebook_2019} text-only translation model.

The Gated Fusion model is a conventional Transformer-based \citep{vaswani_attention_2017} MMT model which takes as input the source text and the image. It contains learnable gating matrices that integrate the multimodal information while allowing for interpretability in terms how the model weighs text and image information to produce its output.

The RMMT method is a Transformer-based retrieval-augmented model which takes as input only the source text. It uses the source text to retrieve contextually relevant images to aid in producing its output. 

For comparison, we also report the performance results achieved by \citet{futeral_tackling_2022}, where they introduced the CoMMuTE framework and their VGAMT model. The VGAMT method is a conventional Transformer-based MMT model that was trained using visually-conditioned masked language modeling (VMLM).

For the Gated Fusion model, we evaluate against the Multi30k test sets as was done in their work. We also evaluate against the Multi30k test sets and newstest test sets in a \ned{non-matching} manner, that is, by associating random images in the Multi30k dataset to each sentence.
For the RMMT model, we use the source text as input and the image retrieval query as done in their work.

\ned{Since we are evaluating against trained models, we perform the respective pre-processing required for the trained models.} For the Gated Fusion and RMMT models, we lowercase, normalize punctuation, tokenize, and perform BPE encoding using the moses tokenizer and subword-nmt. For the RMMT model, we use the lowercased and tokenized source text for the retrieval query as was done in their work. Similarly for the FAIR-WMT19 model, we perform the tokenization and BPE encoding pre-processing steps done by \citet{ng_facebook_2019}.

\begin{table}[t]
\centering

\footnotesize
\begin{tabular}{*{6}{c}}
\toprule
Label & \multicolumn{1}{c}{\scriptsize{CoMMuTE}} & \multicolumn{2}{c}{\scriptsize{Multi30k}} & \multicolumn{2}{c}{\scriptsize{newstest}} \\ \hline

& \multicolumn{1}{c}{} & 2016 & 2017 & 2019 & 2020 \\ \hline

& Score & \multicolumn{4}{c}{BLEU4} \\ \hline

\multicolumn{6}{c}{Multimodal inputs} \\
\scriptsize{Gated Fusion}    & 0.50 & 42.0 & 33.6 & \multicolumn{2}{c}{} \\
\scriptsize{VGAMT}              & 0.59 & 43.3 & 38.3 & \multicolumn{2}{c}{} \\ \hline

\multicolumn{6}{c}{Text inputs only} \\

\scriptsize{FAIR-WMT19}      & 0.50 & 40.7 & 37.7 & 40.6 & 36.2 \\
\scriptsize{RMMT}            & 0.50 & 41.5 & 33.0 &  1.3 &  0.8 \\ \hline

\multicolumn{6}{c}{\ned{Non-matching} inputs} \\
\scriptsize{Gated Fusion}       & 0.50 & 42.0 & 33.6 & 1.3 & 0.6 \\ \bottomrule
\end{tabular}

\caption{BLEU4 scores against the Multi30k test sets and the newstest test sets as well as CoMMuTE scores for English to German translations (en-de). Performance of the Gated Fusion and RMMT models \citep{wu_good_2021} trained against the Multi30k dataset is shown. The results of the FAIR-WMT19 \citep{ng_facebook_2019} text-only model is shown for comparison.}
\label{tab:main}
\end{table}

\section{Results and Discussion}
\label{sec:experiments}

Since both the Gated Fusion and RMMT models are trained solely against the Multi30k dataset, we expect there to be a drop in performance when evaluating against out-of-domain testing sets such as the newstest test sets. As shown in Table \ref{tab:main}, while the two models perform well against the Multi30k test sets, we found the drop in performance against the newstest test sets to be very large, making the two models impractical for real-world usage.

Specifically, for the Gated Fusion model, we found performance to be identical no matter which image was associated with the sentence, which indicates that the model is not be using image information to a significant degree. The fact that the Gated Fusion model learns to ignore image information was discussed by \citet{wu_good_2021}. Furthermore, while the Gated Fusion method performed comparably against the FAIR-WMT19 model on the Multi30k test sets, we see that there is a drastic loss of performance when evaluated against the newstest test sets.

For the RMMT model, we use the source text as the retrieval query for the image retrieval, as was done in the original paper. Similar to the Gated Fusion method, there is a drastic loss of performance when evaluated against the newstest test sets.

We theorize that the primary reasons for this drastic drop in performance are \ned{1) domain mismatch between the image captions in the Multi30k dataset compared to the longer, more complex newstest test sets,} 2) the difference in lengths in the input sentences between the newstest test sets and the Multi30k training set (e.g., the average number of words per example in both the Multi30k training set and the Multi30k test2016 is 12 while it is 34 in newstest2020), 3) the difference in vocabulary between the newstest test sets and the Multi30k training set (e.g., out of approximately 8200 unique words in newstest2020, 5280 words are not in Multi30k training set, while out of approximately 1890 unique words in Multi30k test2016, only 140 words are not in the Multi30k training set), and 4) overfitting on the very small Multi30k training set via both the training process and the BPE pre-processing (29,000 examples).

\thanksnostar{
Opinions, interpretations, conclusions, and recommendations are those of the authors and are not necessarily endorsed by the United States Government. Cleared for public release on 12 Feb 2024. Originator reference number RH-24-125350. Case number AFRL-2024-0794.
}

\section{Conclusion}
\label{sec:conclusion}

Multimodal translation models are typically trained against the Multi30k dataset or a dataset containing relatively few number of captions. 
However, practical machine translation requires that models be able to translate complex sentences with contextual image information.
Using this principle, we introduce an evaluation framework for MMT models that 1) measures their use of visual information to aid in the translation task and 2) measures their ability to translate complex sentences such as done for text-only MT.

We also show that while current models trained against the Multi30k dataset may perform well against the Multi30k testing sets, they can perform poorly against other testing sets used to evaluate text-only models. This calls for a need to evaluate against these testing sets in the MMT domain.

Our results on current MMT models also suggest that MMT models be designed and trained in such a way that they have a baseline high performance on text-only translation. 
For future work, we plan to adapt a text-only MT model to MMT and then fine-tune the MMT model while retaining its text-only translation performance.

\section*{Acknowledgements}

This work is sponsored by the Air Force Research Laboratory
under Air Force contract FA8650-20-D-6207.

\bibliography{zot}

\end{document}